\documentclass[sigconf,nonacm]{acmart}
\setcopyright{none}
\usepackage{algorithm}
\usepackage{algorithmic}
\usepackage{placeins}
\usepackage{multirow}
\usepackage{tabularx}
\usepackage{colortbl}

\newenvironment{treememalgbody}{\begin{algorithmic}[1]}{\end{algorithmic}}
\newcommand{\algio}[2]{\STATE \textbf{#1:} #2}
\title[TreeAgent: Structure-Aware Memory Agent]{TreeAgent: Structure-Aware Memory Agent for Processing \texorpdfstring{\\}{ }Long Hierarchical Data}
\makeatletter
\newcommand{\preprintAuthorNames}{}
\newcommand{\preprintauthor}[4]{%
  \author{#1}#3\email{#4}%
  \g@addto@macro\preprintAuthorNames{%
    \mbox{#1\textsuperscript{\ensuremath{#2}}}\quad}%
}
\newcommand{\preprintauthorbreak}{%
  \g@addto@macro\preprintAuthorNames{\unskip\par}%
}
\newcommand{\preprintaffiliation}[1]{%
  \affiliation{\institution{#1}\city{Hangzhou}\country{China}}%
}
\makeatother

\preprintauthor{Wei Gu}{1,3}{%
  \preprintaffiliation{The State Key Laboratory of Blockchain and Data Security, Zhejiang University}%
  \preprintaffiliation{Alibaba Group}%
}{wei.gu@zju.edu.cn}

\preprintauthor{Weiwei Qi}{1}{%
  \preprintaffiliation{The State Key Laboratory of Blockchain and Data Security, Zhejiang University}%
}{weiweiqi@zju.edu.cn}

\preprintauthor{Jiajun Tang}{1}{%
  \preprintaffiliation{The State Key Laboratory of Blockchain and Data Security, Zhejiang University}%
}{jiajuntang@zju.edu.cn}

\preprintauthor{Tianhang Zheng}{1,2,\dagger}{%
  \preprintaffiliation{The State Key Laboratory of Blockchain and Data Security, Zhejiang University}%
  \preprintaffiliation{Hangzhou High-Tech Zone (Binjiang) Institute of Blockchain and Data Security}%
}{zthzheng@zju.edu.cn}
\preprintauthorbreak

\preprintauthor{Bo Chen}{3}{%
  \preprintaffiliation{Alibaba Group}%
}{wujian.cb@alibaba-inc.com}

\preprintauthor{Zhibo Zhang}{3}{%
  \preprintaffiliation{Alibaba Group}%
}{huaibian.zzb@alibaba-inc.com}

\preprintauthor{Jinfei Liu}{1,2}{%
  \preprintaffiliation{The State Key Laboratory of Blockchain and Data Security, Zhejiang University}%
  \preprintaffiliation{Hangzhou High-Tech Zone (Binjiang) Institute of Blockchain and Data Security}%
}{jinfeiliu@zju.edu.cn}

\preprintauthor{Kui Ren}{1,2}{%
  \preprintaffiliation{The State Key Laboratory of Blockchain and Data Security, Zhejiang University}%
  \preprintaffiliation{Hangzhou High-Tech Zone (Binjiang) Institute of Blockchain and Data Security}%
}{kuiren@zju.edu.cn}

\renewcommand{\shortauthors}{Gu et al.}

\makeatletter
\def\@mkauthors@iii{%
  \hsize=\textwidth
  \global\setbox\mktitle@bx=\vbox{%
    \unvbox\mktitle@bx
    \par\medskip
    \centering
    {\large\normalfont\preprintAuthorNames\unskip\par}%
    \smallskip
    {\normalsize\normalfont
      \textsuperscript{1}The State Key Laboratory of Blockchain and Data Security, Zhejiang University\par
      \textsuperscript{2}Hangzhou High-Tech Zone (Binjiang) Institute of Blockchain and Data Security\par
      \textsuperscript{3}Alibaba Group\par
    }%
    \smallskip
    {\small\normalfont\textsuperscript{\ensuremath{\dagger}}Corresponding author:
      \href{mailto:zthzheng@zju.edu.cn}{zthzheng@zju.edu.cn}\par}%
    \medskip
  }%
}
\makeatother

\begin{document}
\begin{abstract}
Practical Web applications increasingly require agents to interpret long conversations or documents.
As the data length goes beyond the context windows of base large language models (LLMs), recent research proposes memory agents to process the input segment by segment and retain information from previous segments within a bounded memory. However, existing memory agents typically divide the input into flat token segments without preserving its structure or recording its hierarchy, leading to three limitations when processing structured data such as API responses or HTML documents.
First, fixed-length token segmentation can separate the fields belonging to the same structured record, breaking their inherent relationships.
Second, individual segments lack hierarchical information, making their structural context and positions unclear.
Third, successive memory updates may discard the details from earlier necessary sibling records for reasoning on later segments.
To address these limitations, we propose \textbf{TreeAgent}, a structure-aware agent that uses the input hierarchy to guide both segment-wise context processing and access to past information.
Specifically, TreeAgent constructs segments from complete subtrees and groups of adjacent sibling subtrees to preserve local structural relationships.
We provide theoretical guarantees on segment counts to guide token budget selection toward matching the segment count of fixed-length segmentation.
We further attach structural paths to the segments for the agent to identify their ancestors and hierarchical positions. 
We also introduce residual memory links from the segments of low-level children to the segments of high-level parents, enabling direct information flows beyond successive memory updates.
We evaluate TreeAgent with four representative memory agents on eight retrieval and aggregation tasks.
The inputs are derived from real-world application data in JSON, HTML, YAML, and JavaScript and range from 8K to 1M tokens.
TreeAgent improves the average score across the four datasets for all four agents, with 3.61\% performance gain on average.
Our code is available at \url{https://github.com/ZJU-LLM-Safety/TreeAgent}.
\end{abstract}

\keywords{Memory agents, hierarchical data, structure-aware processing, long-context reasoning}
\maketitle

\section{Introduction}
Large language model (LLM) agents have made significant advances across diverse domains in recent years \citep{yao2023reactsynergizingreasoningacting,openai2025_chatgpt_agent,lihermes,wang2026agentsnarelearningdelaydivert,zeng2026trace}.
As these agents process increasingly long inputs, the input length can exceed the context windows of their underlying models \citep{anthropic2025_context_engineering}.
To address this over-length issue, recent research has introduced memory agents,
which process the input segment by segment and continually update a bounded memory to retain relevant information for subsequent reasoning.
Memory agents have demonstrated strong performance on question-answering tasks involving documents far longer than the LLMs' native context windows \citep{yu2025memagentreshapinglongcontextllm,wang2026infmemlearningsystem2memorycontrol}.

However, when processing long structured data such as API responses and HTML documents in Web applications\citep{zhong2025complexfuncbenchexploringmultistepconstrained,chen2021websrc}, existing memory agents exhibit the following three limitations.
First, existing methods typically split the input into fixed-length token segments, which can separate related fields within the same record, so associations already represented in the original structure may need to be reconstructed across segments.
Second, existing memory agents process each segment without an explicit indication of its position in the input hierarchy. Without this information, the agent may form incorrect associations between segments, leading to flawed subsequent reasoning~\citep{chen2021websrc,li2022markuplm}.
Third, successive updates and compression of bounded memory may discard information needed for subsequent reasoning across segments and records~\citep{wang2026infmemlearningsystem2memorycontrol,shi2025lookbackreasonforward}.

To address these limitations, we propose \textbf{TreeAgent}, a structure-aware agent that uses the input hierarchy to guide both segment-wise processing and access to historic information.
TreeAgent preserves complete subtrees or groups of adjacent sibling subtrees within a segment whenever the token budget permits, making their local structural relationships available within a single memory update.
Starting from the entire input tree, TreeAgent recursively applies this segmentation strategy to its subtrees whenever the current tree exceeds the token budget.
Grouping adjacent sibling subtrees reduces the number of small segments that would otherwise be emitted separately.
However, preserving subtree boundaries can still leave unused capacity, producing more segments than fixed-length segmentation at the same token budget.
We analyze the optimality of greedy sibling grouping and establish the budget needed to compensate for unused capacity within a sibling sequence.
Based on the theoretical guarantees derived from this analysis, we adjust the token budget to keep the segment count close to that of fixed-length segmentation.
Each segment is augmented with a structural path that explicitly links its local content to the enclosing structure, enabling the agent to interpret the current input within its hierarchical context for reasoning across related segments.

TreeAgent further introduces residual memory links to bring fine-grained information from historic memory into reasoning at higher levels. During segment-wise processing, TreeAgent saves the memory after each segment and uses the parent–child relationships in the original input to link historic memories associated with previously processed children to the final segment of their parent. 
When processing this segment, the agent retrieves information relevant to the question from the linked memory states and uses it together with the current segment and its structural path to update memory. 
In this way, relevant details retained in historic memory can bypass intermediate memory updates and be integrated directly during processing at higher levels of the hierarchy, supporting comparison and aggregation across records.

We evaluate TreeAgent with four representative memory agents: MemAgent, InfMem, MemReread, and ReMemR1.
The evaluation uses structured inputs derived from real-world application data in JSON, HTML, YAML, and JavaScript.
Input lengths range from 8K to 1M tokens.
Across these formats, we assess eight retrieval and aggregation tasks that require locating values and reasoning about their structural relationships.
The tasks include retrieving multiple fields from one record, retrieving fields across records, counting matching records, identifying frequent values, and finding numerical extrema.
For each agent, we compare TreeAgent with its corresponding fixed-length segmentation baseline while keeping the model parameters and existing memory update and answering procedures unchanged.
The baselines retain each agent's native retrieval, rereading, or callback operations when available, allowing the comparisons to assess the benefit of TreeAgent within the existing workflows.
TreeAgent improves each agent's average score across the four datasets, with a mean gain of 3.61\% over the corresponding baselines.
Sequential ablations show that structural paths and residual memory provide further improvements beyond structural segmentation alone.

Our contributions are summarized as follows:
\begin{enumerate}
    \item {\emergencystretch=0.5em
We propose TreeAgent, a structure-aware agent that partitions long
hierarchical inputs into subtree sections and equips each section with
an explicit structural path, preserving local structural relationships
and hierarchical context throughout recurrent memory processing.\par}

    \item We introduce residual links in TreeAgent to directly transmit
historic evidence relevant to the question from low-level child subtrees
to high-level parents, mitigating information loss across successive
memory updates.

    \item We evaluate TreeAgent on retrieval and aggregation tasks over long structured inputs derived from real-world application data. Comparative experiments with four representative memory agents and component ablations demonstrate its effectiveness across diverse structured formats.
\end{enumerate}

\section{Preliminaries}

\paragraph{Recurrent Memory Agents.}
Given an input $x$ and a query $q$, an ordered sequence of segments $s_1,\ldots,s_T$ is constructed from the input, where $T$ denotes the number of segments.
Segmentation determines which parts of the input are read together at each step.
The agent processes these segments sequentially and updates its memory after each one.
At step $t$, the agent receives an observation $o_t$, comprising the current segment $s_t$ and any additional context, and combines it with the query $q$ and previous memory $m_{t-1}$ to produce the updated memory state $m_t$.
Let $\mathcal A$ denote the agent workflow and $\theta$ the model parameters.
The memory update is expressed as
\begin{equation}
m_t=\mathcal U_{\mathcal A,\theta}(q,m_{t-1},o_t).
\end{equation}
Different memory agents instantiate this interface through their respective workflows, with recurrent memory constrained by the corresponding agent's memory budget \citep{wang2026infmemlearningsystem2memorycontrol,shi2025lookbackreasonforward}.

\paragraph{Fixed-length Segmentation.}
Given the agent's tokenizer $\tau$ and a segment content budget $C$, fixed-length segmentation partitions the token sequence $\tau(x)$ into consecutive segments in input order.
Each segment contains $C$ tokens, except possibly the last.
We refer to the baseline using this segmentation strategy as Flat.
The number of segments it produces for input $x$ at budget $C$ is
\begin{equation}
T_{\mathrm{flat}}(x,C)
=
\left\lceil\frac{|\tau(x)|}{C}\right\rceil,
\end{equation}
where $|\tau(x)|$ denotes the number of input tokens.
The budget $C$ applies only to segment content, excluding the query, memory, and additional context.
For a given agent, $\tau$ is fixed and its dependence is omitted from the segment count notation.

Let $c_t^{\mathcal A}$ denote any additional context supplied by the existing agent workflow at step $t$, with $c_t^{\mathcal A}=\emptyset$ when none is supplied.
Under Flat, the current segment and this additional context together form the observation supplied to the agent's memory update:
\begin{equation}
o_t^{\mathrm{Flat}}=(s_t,c_t^{\mathcal A}).
\end{equation}

\section{TreeAgent: A Structure-Aware Agent for Reasoning over Hierarchical Data}

\begin{figure*}[t]
\centering
\includegraphics[width=\textwidth]{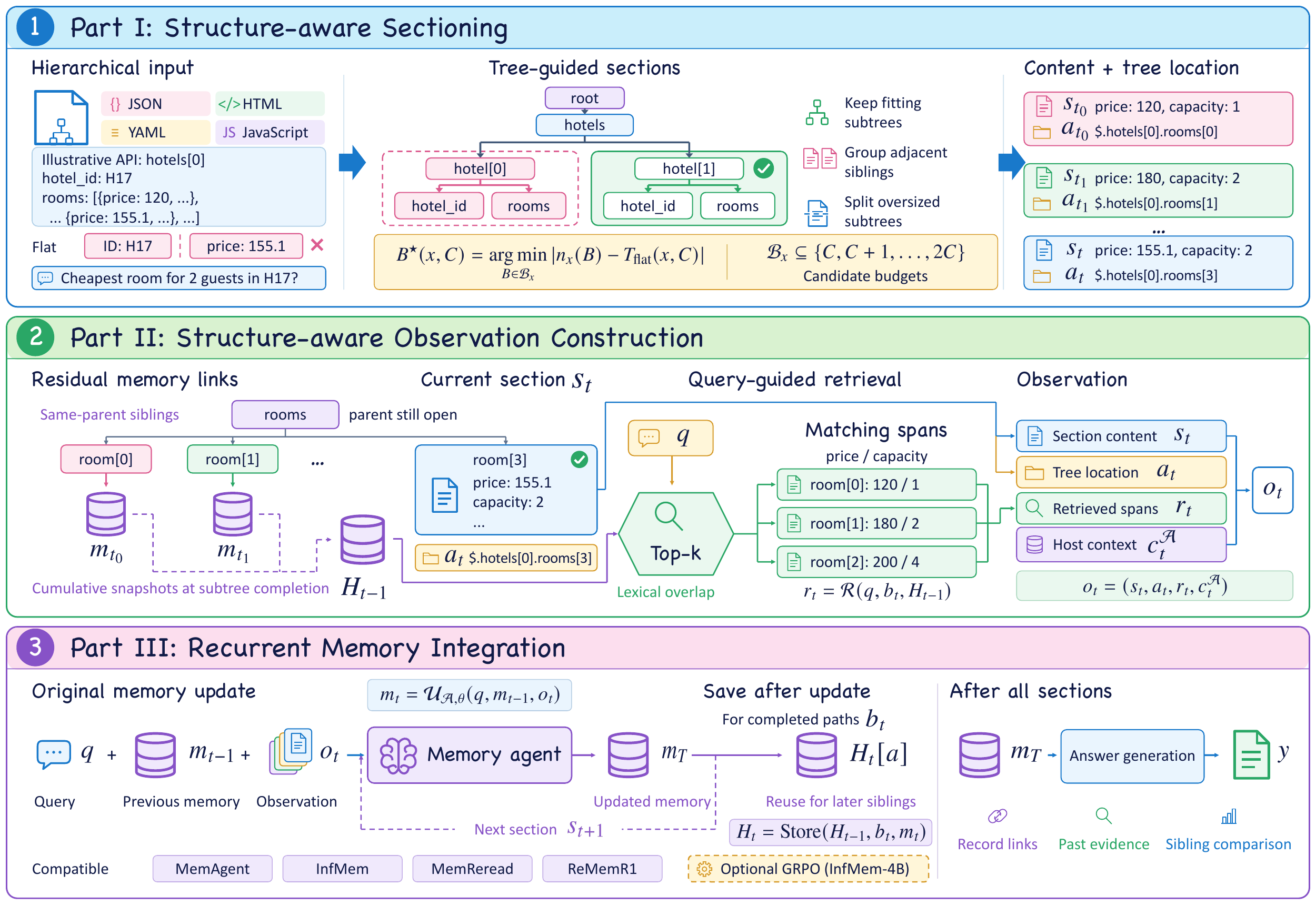}
\caption{Overview of TreeAgent, showing how an observation is constructed and used in a segment-wise memory update. Each observation $o_t$ combines the current structural section, its path, and spans relevant to the query retrieved from historic memory states through residual memory links. The memory-agent block represents the agent's original workflow, including its internal actions and stopping behavior. After processing each section, TreeAgent saves the updated memory. Parent--child relations determine which saved states are linked and accessed by later updates.}
\label{fig:treeagent_overview}
\Description{Diagram of structural sectioning, paths, residual retrieval, and recurrent memory updates.}
\end{figure*}

TreeAgent uses the input hierarchy to organize both the content of each observation and access to historic evidence (Figure~\ref{fig:treeagent_overview}).
Structural sectioning keeps fitting subtrees intact and groups adjacent siblings to control the section count.
Structural paths identify where each section belongs, while residual memory links make memories associated with processed children available when the final segment of their parent is processed.
Together, these components supply local content, its hierarchical context, and relevant historic evidence to the agent's existing memory update procedure.

\subsection{Sectioning Hierarchical Inputs}
\label{sec:structural-sectioning}

TreeAgent favors complete subtrees when constructing sections so that fields within each retained subtree are available together during a single memory update.

\paragraph{Tree Representation.}
Let $\mathcal{T}(x)=(V,E,r)$ denote the tree representing the hierarchical structure of $x$, where $r$ is the root, $E$ specifies parent--child relations, and siblings retain their original document order.
For each node $v\in V$, $\operatorname{tok}(v)$ denotes the token count of the complete serialization of the subtree rooted at $v$.
For API responses represented in JSON, objects and arrays form internal nodes, and scalar fields form leaves.
The same abstraction applies to HTML, YAML, and static JavaScript when their parsers expose corresponding hierarchical relations.

A section may contain a single subtree rooted at a node $v$, which we call a \emph{subtree section}.
To better utilize the section budget, adjacent sibling subtrees may also be grouped into a \emph{sibling section}.

\paragraph{Section Construction.}
Algorithm~\ref{alg:tree} constructs an ordered section list $\mathcal{S}_B(x)$ by retaining subtrees that fit within budget $B$.
For a larger internal node, it groups adjacent child subtrees in document order while their combined serialization fits within $B$, recursively partitioning any child that exceeds the budget.
Oversized leaves are split into consecutive text spans within $B$ at boundaries permitted by the input format, except those whose syntax must remain intact.
The indivisible leaves are retained even when they exceed $B$.

\begin{algorithm}[t]
\caption{Sectioning Hierarchical Inputs under a Token Budget}
\label{alg:tree}
\begin{treememalgbody}
\algio{Procedure}{\textsc{Partition}$(v,B)$}
\algio{Output}{ordered section list $K$}
\STATE \textbf{if} $\operatorname{tok}(v)\leq B$ or $\operatorname{atomic}(v)$ \textbf{then return} $[v]$
\STATE \textbf{if} $\operatorname{leaf}(v)$ \textbf{then return} \textsc{SplitText}$(v,B)$
\STATE $K \leftarrow [\,],\quad P \leftarrow [\,]$
\FOR{$u \in \operatorname{children}(v)$ in document order}
\IF{$\operatorname{tok}(u)>B$}
\STATE \textsc{Flush}$(K,P);\quad K \leftarrow K \oplus \textsc{Partition}(u,B)$
\ELSE
\STATE \textbf{if} $\operatorname{tok}(P \oplus [u])>B$ \textbf{then} \textsc{Flush}$(K,P)$
\STATE $P \leftarrow P \oplus [u]$
\ENDIF
\ENDFOR
\STATE \textsc{Flush}$(K,P)$
\RETURN $K$
\end{treememalgbody}
\par\smallskip
\noindent\textit{Notation.} $\operatorname{atomic}$ identifies indivisible leaves; \textsc{SplitText} splits leaves at format-permitted boundaries within $B$. \textsc{Flush}$(K,P)$ appends nonempty $P$ to $K$ and clears $P$. $\operatorname{tok}$ counts serialized tokens for subtrees or sibling lists.
\end{algorithm}

\paragraph{Structure Preservation.}
Let $K_B(v)$ be the number of sections to which the source content of subtree $v$ is assigned.
Algorithm~\ref{alg:tree} preserves document order and satisfies
\begin{equation}
\operatorname{tok}(v)\le B \quad\Longrightarrow\quad K_B(v)=1.
\label{eq:subtree-preservation}
\end{equation}
If an ancestor of $v$ is retained, its section contains all of $v$.
Otherwise, recursion reaches $v$ and retains it either alone or within a sibling group.
Thus, fields within a subtree that fits the budget appear together in one observation, making their local relations available within a single memory update.

\paragraph{Grouping Adjacent Siblings.}
Emitting each fitting subtree separately can leave much of the section budget unused.
Grouping adjacent siblings uses this capacity while retaining their common parent context and document order.
For the section-count analysis, consider an ordered tree whose leaf units have positive integer lengths at most $B$.
Let $w(v)$ be the sum of the lengths of the leaves in its subtree; subtree and sibling-group lengths are additive.
A feasible partition covers these units exactly once, in order, using subtree or sibling sections of length at most $B$.
Under this model, greedy grouping minimizes the section count among feasible partitions.

Consider a sequence of fitting siblings with lengths $w_1,\ldots,w_m$.
Let $D(i)$ be the minimum number of sections needed for the suffix beginning at $i$ without splitting its subtrees, with $D(m+1)=0$.
Deleting a prefix cannot increase this minimum, so $D(i)$ is nonincreasing in $i$.
If the longest prefix fitting within $B$ ends at $j$, then
\begin{equation}
D(1)=1+\min_{1\le k\le j}D(k+1)=1+D(j+1).
\label{eq:sibling-optimality}
\end{equation}
Greedy grouping chooses this prefix and repeats the same optimal choice on the remaining suffix.

Under this model, optimality extends to the tree.
Any fitting subtree split by a feasible partition can be replaced by one subtree section without increasing the count: its proper descendants cannot share a sibling section with nodes outside it.
An oversized child separates the siblings before and after it into independent packing sequences and is partitioned recursively.
Applying Equation~\eqref{eq:sibling-optimality} to each sequence and induction to each oversized child gives
\begin{equation}
N_B=\min_{\mathcal P\in\mathcal A_B}|\mathcal P|,
\label{eq:tree-section-optimality}
\end{equation}
where $N_B$ is the count produced by Algorithm~\ref{alg:tree} using lengths $w$, and $\mathcal A_B$ is the set of feasible partitions.
Thus, retaining fitting subtrees and greedily grouping siblings jointly achieve the minimum section count under the structural constraints.

\paragraph{Controlling the Section Count.}
The budget $B$ determines both which subtrees can remain intact and how many sections the agent processes.
Keeping sibling subtrees intact can leave unused capacity even with optimal grouping.
Flat fills each section with $C$ tokens, except possibly the last one.
We therefore seek a budget that compensates for this packing loss, allowing a sibling sequence to use no more sections than Flat at budget $C$.
The following result establishes $2C-1$ as the smallest integer budget with the following guarantee.

\begin{proposition}[Tight budget for sibling packing]
\label{prop:tight-sibling-budget}
Fix a positive integer $C$.
Let $N(B)$ denote the number of sections produced by greedy sibling packing at budget $B$.
For every nonempty ordered sequence of indivisible sibling subtrees with positive integer lengths $w_i\le 2C-1$ and total length $W=\sum_iw_i$, greedy packing satisfies
\begin{equation}
N(2C-1)\le\left\lceil\frac{W}{C}\right\rceil.
\label{eq:tight-sibling-budget}
\end{equation}
Moreover, $2C-1$ is the smallest integer budget $B\ge C$ that guarantees $N(B)\le\lceil W/C\rceil$ for every such sequence with $w_i\le B$.
\end{proposition}

\begin{proof}
Set $B=2C-1$, and let $L_1,\ldots,L_N$ be the resulting section lengths.
The claim is immediate for $N=1$.
For each adjacent pair, the first subtree of the next section did not fit in the preceding section.
Hence $L_i+L_{i+1}>B$, and integer lengths imply $L_i+L_{i+1}\ge2C$.
Summing over the $N-1$ boundaries gives
\begin{equation}
2C(N-1)\le 2W-L_1-L_N\le 2W-2.
\label{eq:packing-boundary-sum}
\end{equation}
Thus $W\ge C(N-1)+1$, which yields Equation~\eqref{eq:tight-sibling-budget}.

For tightness, take $C\ge2$ and any integer $B$ with $C\le B\le2C-2$.
Use $C$ repetitions of the pair $(C-1,C)$.
Each subtree fits, but every adjacent pair has length $2C-1>B$, so every subtree requires a separate section.
Consequently,
\begin{equation}
N(B)=2C>2C-1=\left\lceil\frac{C(2C-1)}{C}\right\rceil.
\label{eq:packing-tightness}
\end{equation}
No smaller budget in the range $B\ge C$ suffices; for $C=1$, the minimum is already $B=1$.
\end{proof}

Proposition~\ref{prop:tight-sibling-budget} shows that budget $2C-1$ suffices to preserve each fitting sibling subtree while using no more sections than Flat at budget $C$ for the same sequence.
Its tightness gives a theoretical basis for choosing $2C-1$ as the upper endpoint of the budget search.
Recursive partitioning produces sibling sequences that are packed separately.
Applying the same boundary argument within these sequences yields a bound for the complete tree.
\begin{corollary}
\label{cor:recursive-section-count}
Under the additive length model, let $W=w(r)$ and let $F_B$ count the boundaries between the sibling sequences packed separately by recursion, treating an intact root as one group.
At $B=2C-1$, the complete tree satisfies
\begin{equation}
N_{2C-1}\le\left\lceil\frac{W}{C}\right\rceil+F_{2C-1}.
\label{eq:recursive-section-count}
\end{equation}
\end{corollary}
\begin{proof}
For $N_B=1$ the claim is immediate.
Otherwise, each of the $N_B-1-F_B$ within-group boundaries contributes at least $2C$ to the sum of adjacent section lengths.
Their sum is at most $2W-2$, since the first and last sections each contribute at most once and all others at most twice.
Therefore $W\ge C(N_B-1-F_B)+1$.
\end{proof}

Based on the sibling packing bound, we search within $[C,2C-1]$, using Flat's budget as the lower endpoint and the tight sibling packing threshold as the upper endpoint.
Let $n_x(B)=|\mathcal{S}_B(x)|$ denote the resulting section count and $\mathcal{B}_x\subseteq\{C,C+1,\ldots,2C-1\}$ the candidate budgets considered by the search.
We select the candidate whose section count is closest to that of Flat:
\begin{equation}
B^\star(x,C)
=
\arg\min_{B\in\mathcal{B}_x}
\left|n_x(B)-T_{\mathrm{flat}}(x,C)\right|.
\label{eq:budget-selection}
\end{equation}

{\emergencystretch=1em
For each candidate budget, we construct sections using the agent's
 tokenizer and compute $n_x(B)$.
For equal count differences, we prefer a candidate whose count is not
 below $T_{\mathrm{flat}}(x,C)$, then the smaller budget.
The selected budget determines how many sections the agent processes:
$T_{\mathrm{tree}}(x,C)=n_x(B^\star(x,C))$.\par}

\subsection{Structural Paths}
Structural segmentation keeps related content together whenever possible, while structural paths identify its position in the overall hierarchy.
TreeAgent pairs each segment $s_t$ with a structural path $a_t$ derived from the original input tree, allowing the agent to interpret local content within its hierarchical context.

The path follows the ancestor chain from the root to the target node.
For a nonroot node $v$ with parent $p$, let $\lambda(v)$ denote its identifier within $p$, determined by its name or position in the input structure and distinguishing it from the other children of $p$.
The node path is defined recursively as
\begin{equation}
\begin{aligned}
\pi(r)&=\epsilon,\\
\pi(v)&=\pi(p)\mathbin{\Vert}\langle\lambda(v)\rangle,
\end{aligned}
\end{equation}
where $r$ is the root, $\epsilon$ denotes the empty root path, and $\Vert$ denotes sequence concatenation.
Appending each child's identifier to its parent's path preserves the ancestor chain from the root.

TreeAgent constructs the path of each segment according to the structural units it covers.
For a segment formed from a single subtree rooted at $v$, we use $a_t=\pi(v)$.
For a segment formed by grouping consecutive child subtrees $u_i,\ldots,u_j$ of a parent $p$, with $i<j$, we use $a_t=(\pi(p),[i,j])$, combining the parent's path with the inclusive range of covered children in input order.
When an oversized leaf is split into text spans, each resulting segment retains the path of that leaf.
These rules give
\begin{equation}
a_t=
\begin{cases}
\pi(v), & \text{subtree rooted at }v,\\
(\pi(p),[i,j]), & \text{siblings }u_i,\ldots,u_j,\ i<j,\\
\pi(v), & \text{text span from leaf }v.
\end{cases}
\end{equation}
For grouped subtrees, the parent path identifies their shared hierarchical context, while the child range specifies the portion covered by the segment.
For a split leaf, its path preserves the structural location of the source content.
Because paths are derived from the original input tree, a segment retains this location information even when it does not contain the full enclosing structure.

\subsection{Residual Memory Links}
Combining information across segments can require historic evidence that is no longer retained in the current memory.
TreeAgent retrieves relevant evidence from previously saved memory states through residual memory links organized by the input hierarchy.
After processing each segment $s_t$, TreeAgent saves the updated memory $m_t$ under the segment index $t$.
For a node $v\in V$, let $\ell(v)$ denote the index of the final segment containing content from its subtree.
TreeAgent associates $v$ with the saved memory state $m_{\ell(v)}$, which contains the agent's cumulative memory after that segment has been processed.
For a leaf split across multiple text spans, $\ell(v)$ identifies the segment containing its final span.

TreeAgent constructs residual links from the input tree and the boundaries between consecutive segments.
For each parent $p$ and child $v$, TreeAgent links the memory associated with $v$ to the final segment of $p$, provided that the source memory is available before the target segment is processed.
The resulting links are
\begin{equation}
\begin{aligned}
\mathcal E_{\mathrm{res}}
&=\{\,(\ell(v),\ell(p))\mid p\in V,\\
&\qquad v\in\operatorname{children}(p),\\
&\qquad \ell(v)<\ell(p)\,\}.
\end{aligned}
\end{equation}
A link $(i,t)$ allows the agent to access the saved memory $m_i$ when processing segment $s_t$.
The strict inequality ensures that the source state has already been saved; the root can also serve as a target parent.
Thus, the links can be determined from the tree and segment boundaries before processing, while their source memories become available as processing proceeds.

For the current segment $s_t$, the incoming links identify the accessible memory indices.
These indices specify the saved memory states available for retrieval during the current update:
\begin{equation}
\mathcal L_t=\{\,i\mid(i,t)\in\mathcal E_{\mathrm{res}}\,\}.
\end{equation}
When the final segments of multiple parents coincide, their valid links are combined.
Each source index appears only once in $\mathcal L_t$, so multiple links to the same saved memory state do not duplicate that state in the retrieval candidates.

Within these linked memory states, the question guides the selection of relevant evidence.
TreeAgent splits the states indexed by $\mathcal L_t$ into short text spans and ranks them by lexical overlap with the question.
It discards spans with zero overlap and selects at most $k$ spans with the highest scores, truncating each to a specified token limit.
The selected spans are concatenated to form the retrieved context $r_t$.
If $\mathcal L_t=\emptyset$ or no span matches the question, $r_t=\emptyset$.
The hierarchy thus determines where to retrieve from, and the question determines what to retrieve.

\paragraph{Integration with Memory Agents.}
These components define the observation supplied to the memory agent: the segment provides local content, its path identifies the structural location, and retrieved spans supply historic evidence.
TreeAgent combines them with the context $c_t^{\mathcal{A}}$ supplied by the original workflow, retaining the agent's memory update and answer procedures.
Let $H_{t-1}$ denote the set of memory states saved before step $t$, each paired with its segment index, with $H_0=\emptyset$.
Retrieval uses only these historic states; the updated memory is saved after the current update:
\begin{equation}
\begin{aligned}
r_t &= \mathcal{R}(q,\mathcal L_t,H_{t-1}),\\
o_t &= (s_t,a_t,r_t,c_t^{\mathcal{A}}),\\
m_t &= \mathcal{U}_{\mathcal{A},\theta}(q,m_{t-1},o_t),\\
H_t &= H_{t-1}\cup\{(t,m_t)\}.
\end{aligned}
\end{equation}
The update combines information accumulated in the recurrent memory $m_{t-1}$ with historic evidence retrieved through $r_t$.
Residual memory links therefore allow evidence in saved states to contribute again without having to survive every intervening memory update.

\section{Experiments}

We evaluate whether TreeAgent improves existing memory agents across structured input formats, context lengths, and task types without additional training.
Component ablations and computational measurements further assess the contributions of its structural components and the associated inference costs.

\subsection{Evaluation Data and Tasks}
\label{sec:evaluation-data}

We collect publicly available structured data from real-world applications, including API responses, web pages \citep{chen2021websrc}, interface specifications \citep{apisguru_openapi_directory}, and static structured data from open-source repositories.
The data cover four formats: JSON, HTML, YAML, and JavaScript.
The JSON dataset contains API responses from USGS \citep{usgs_earthquake_catalog}, NVD \citep{nist_nvd}, NIH \citep{nih_reporter_api,nih_clinicaltrials_api}, GBIF \citep{gbif_api}, Crossref \citep{crossref_rest_api}, Contracts Finder \citep{contracts_finder}, and USAspending \citep{usaspending_api}, covering earthquake events, vulnerabilities, research funding and clinical trials, biodiversity, scholarly metadata, and public procurement and spending.
The HTML data include web pages and DOM records from sources such as WebSRC, while the YAML data include OpenAPI specifications from sources such as APIs.guru.
The JavaScript data include object and array initializers extracted from open-source repositories \citep{eslint_repository,webpack_repository,threejs_repository,plotlyjs_repository,babel_repository,lodash_repository}, representing static structured state in applications.
Inputs retain the syntax and hierarchical structure of their respective formats.

The evaluation comprises eight tasks that range from retrieval within a single record to retrieval across multiple records and aggregation within a specified scope.
Three tasks retrieve one, two, or four fields from a single specified record, testing whether agents associate the requested fields with the correct entity.
Two further tasks retrieve the same field from two or four specified records, requiring agents to distinguish entities and collect the corresponding information.
The remaining three tasks require aggregation over records within the scope specified by the question: counting records that satisfy a condition, identifying the most frequent values, and finding a numerical extremum.

The evaluation uses four datasets, one per format, covering eight input lengths from 8K to 1M tokens.
Natural language questions are formulated from the records in each input according to the eight task definitions, with 50 test examples per task and input length.

\subsection{Experimental Setup}

\paragraph{Memory agents.}
We evaluate MemAgent-7B \citep{yu2025memagentreshapinglongcontextllm}, InfMem-4B \citep{wang2026infmemlearningsystem2memorycontrol}, MemReread-4B \citep{ji2026memreread}, and ReMemR1-7B \citep{shi2025lookbackreasonforward}.
We integrate TreeAgent into their inference procedures without additional training.
Within each agent, Flat and TreeAgent share model parameters and the original memory update and answer procedures.

\paragraph{Evaluation protocol.}
Within each agent, Flat and TreeAgent share the test examples and evaluation protocol.
Flat uses a segment budget of 5,000 tokens, and TreeAgent adjusts its token budget to approximate Flat's segment count using the procedure in Section~\ref{sec:structural-sectioning}.

\paragraph{Metrics.}
We report mean answer scores.
Single-value and numerical answers receive a score of one if correct and zero otherwise.
For multi-value answers, the score is the number of correctly matched values divided by the larger of the predicted and reference list lengths, with answer order enforced when required by the question.
Dataset scores average equally over tasks and input lengths; overall scores further average equally over datasets and agents.

\subsection{Main Results}

\begin{table*}[t]
\centering
\caption{Mean answer scores (\%) for four memory agents, averaged over eight tasks and eight context lengths (8K--1M). Tree, Tree+Path, and TreeAgent successively add structural sectioning, structural paths, and residual memory links with span retrieval to Flat. AVG. averages the four datasets; the final group averages the four agents. Bold marks the best score within each group and column.}
\label{tab:main_results}
\label{tab:component_ablation}
\begingroup
\normalsize
\setlength{\tabcolsep}{4pt}
\renewcommand{\arraystretch}{1.12}
\definecolor{treeagenttablehighlight}{RGB}{252,240,239}
\begin{tabularx}{\textwidth}{ll*{5}{>{\raggedleft\arraybackslash}X}}
\toprule
\textbf{Memory agent} & \textbf{Method} & \textbf{JSON} & \textbf{HTML} & \textbf{YAML} & \textbf{JavaScript} & \textbf{AVG.} \\
\midrule
\multirow{4}{*}{\textbf{MemAgent-7B}} & Flat & 35.48 & 34.96 & 40.68 & \textbf{50.44} & 40.39 \\
 & Tree & 36.75 & 35.10 & 40.85 & 49.30 & 40.50 \\
 & Tree+Path & 36.94 & 35.22 & 40.62 & 49.20 & 40.50 \\
\rowcolor{treeagenttablehighlight}
\cellcolor{white} & \textbf{TreeAgent} & \textbf{37.28} & \textbf{35.29} & \textbf{42.00} & 50.38 & \textbf{41.24} \\
\cmidrule(lr){1-7}
\multirow{4}{*}{\textbf{InfMem-4B}} & Flat & 57.67 & 38.65 & 65.01 & 63.51 & 56.21 \\
 & Tree & 57.75 & 42.95 & 68.10 & 67.25 & 59.01 \\
 & Tree+Path & 58.15 & 43.60 & 67.95 & 68.45 & 59.54 \\
\rowcolor{treeagenttablehighlight}
\cellcolor{white} & \textbf{TreeAgent} & \textbf{65.12} & \textbf{44.56} & \textbf{68.74} & \textbf{69.04} & \textbf{61.86} \\
\cmidrule(lr){1-7}
\multirow{4}{*}{\textbf{MemReread-4B}} & Flat & 24.07 & \textbf{27.46} & 46.07 & 38.85 & 34.11 \\
 & Tree & 37.11 & 27.30 & 50.05 & 38.70 & 38.29 \\
 & Tree+Path & 38.28 & 27.05 & 50.80 & 39.55 & 38.92 \\
\rowcolor{treeagenttablehighlight}
\cellcolor{white} & \textbf{TreeAgent} & \textbf{39.57} & 27.01 & \textbf{54.41} & \textbf{40.87} & \textbf{40.46} \\
\cmidrule(lr){1-7}
\multirow{4}{*}{\textbf{ReMemR1-7B}} & Flat & 17.39 & 21.96 & 24.22 & 45.49 & 27.26 \\
 & Tree & 18.26 & 22.20 & 24.55 & 45.30 & 27.58 \\
 & Tree+Path & 18.60 & 22.30 & 24.90 & 45.55 & 27.84 \\
\rowcolor{treeagenttablehighlight}
\cellcolor{white} & \textbf{TreeAgent} & \textbf{19.03} & \textbf{22.47} & \textbf{27.35} & \textbf{46.52} & \textbf{28.84} \\
\midrule
\multirow{4}{*}{\textbf{Four-agent mean}} & Flat & 33.65 & 30.76 & 44.00 & 49.57 & 39.49 \\
 & Tree & 37.47 & 31.89 & 45.89 & 50.14 & 41.35 \\
 & Tree+Path & 37.99 & 32.04 & 46.07 & 50.69 & 41.70 \\
\rowcolor{treeagenttablehighlight}
\cellcolor{white} & \textbf{TreeAgent} & \textbf{40.25} & \textbf{32.33} & \textbf{48.13} & \textbf{51.70} & \textbf{43.10} \\
\bottomrule
\end{tabularx}
\endgroup
\end{table*}

\paragraph{Overall performance.}
Table~\ref{tab:main_results} shows that TreeAgent increases the overall mean answer score from 39.49\% to 43.10\%, yielding a gain of 3.61\%.
All four memory agents improve when averaged across the four datasets.
InfMem-4B and MemReread-4B obtain gains of 5.65\% and 6.35\%, respectively, while MemAgent-7B and ReMemR1-7B improve by 0.85\% and 1.58\%.
TreeAgent outperforms Flat in 14 of the 16 comparisons across agents and datasets.
Averaging over agents, gains are positive on all four datasets.
These results show improvements across agents and input formats, with substantial variation in their magnitude across agents.

\begin{figure*}[t]
\centering
\includegraphics[width=\textwidth]{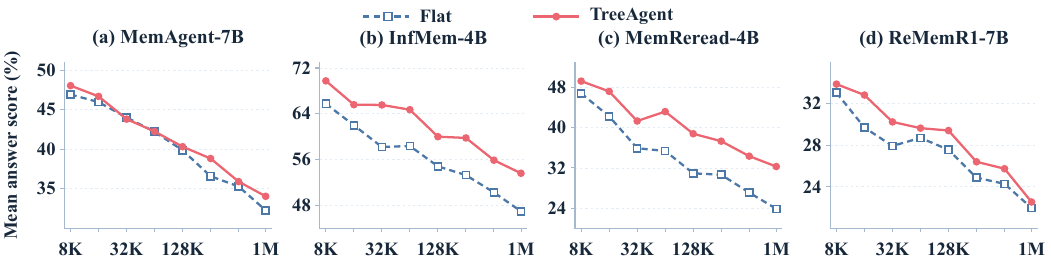}
\caption{Mean answer scores (\%) of Flat and TreeAgent as context length increases from 8K to 1M tokens. Each panel shows one memory agent with fixed model parameters. At each length, scores are averaged over eight tasks with 50 examples per task, and then averaged equally across the four datasets.}
\label{fig:training_free_length_gains}
\Description{Four panels compare Flat and TreeAgent scores across context lengths for the four memory agents.}
\end{figure*}

\paragraph{Performance across context lengths.}
Figure~\ref{fig:training_free_length_gains} reports mean answer scores as inputs grow from 8K to 1M tokens.
Both methods generally score lower on longer inputs.
Nevertheless, TreeAgent outperforms Flat in 31 of the 32 comparisons across agents and context lengths, including all four agents at 1M tokens.
For MemReread-4B, the gain increases from 2.43\% at 8K to 8.40\% at 1M.

\begin{figure*}[t]
\centering
\includegraphics[width=\textwidth]{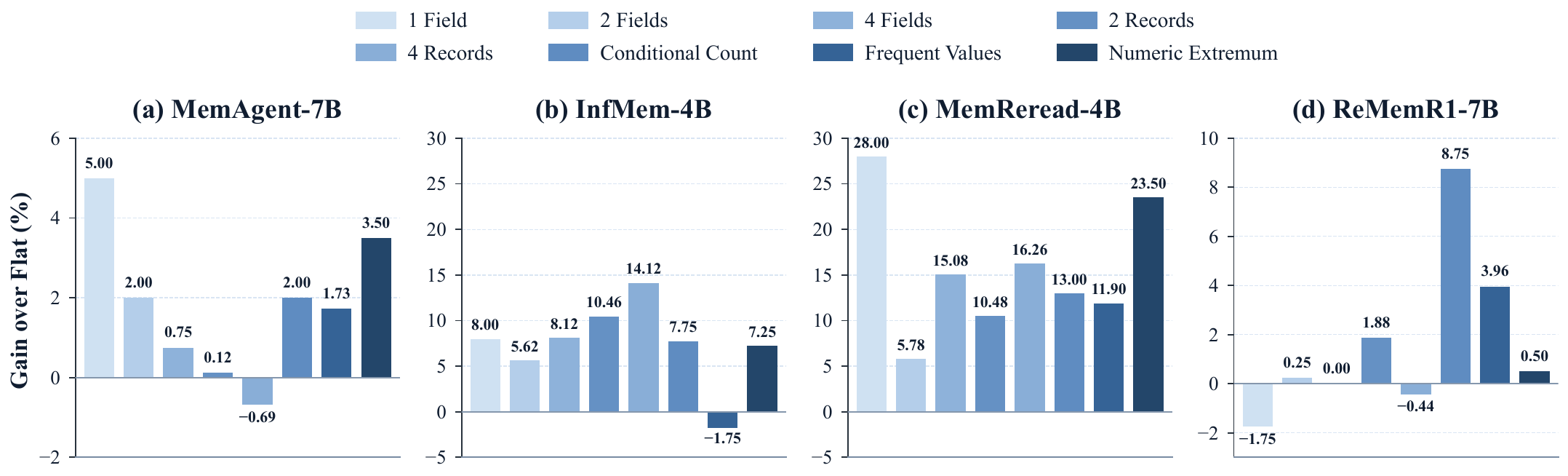}
\caption{Answer score gains of TreeAgent over Flat across eight tasks on the JSON evaluation set (TreeAgent $-$ Flat, \%). Gains are averaged equally over eight context lengths from 8K to 1M tokens, with 50 test examples per task and context length. Panels use different y-axis ranges.}
\label{fig:task_gains}
\Description{Four panels show TreeAgent minus Flat answer scores for eight JSON tasks. Colors distinguish retrieval of one, two, or four fields, retrieval from two or four records, conditional counting, frequent values, and numerical extrema.}
\end{figure*}

\paragraph{Performance across task types.}
We use the JSON evaluation set to examine how the overall improvements are distributed across the eight tasks.
Figure~\ref{fig:task_gains} reports TreeAgent's gains over Flat for each memory agent, with each result averaged equally over the eight context lengths.
Averaging equally across the four agents, TreeAgent improves all eight tasks, with gains ranging from 3.41\% to 9.81\%.
The mean gains for retrieval within a record, retrieval across records, and aggregation are 6.40\%, 6.53\%, and 6.84\%, respectively.
These comparable gains show that the benefits extend from field retrieval within individual records to tasks requiring information collection and aggregation across records.

InfMem-4B gains 10.46\% and 14.12\% on retrieval from two and four records, respectively, exceeding its gains on all three tasks involving retrieval within a single record.
MemReread-4B also obtains substantial gains of 10.48\% and 16.26\% on these two tasks.
For both agents, the improvement is larger when retrieving from four records than from two.
These results demonstrate benefits when agents must collect information from several records, consistent with TreeAgent's design objective of organizing current inputs while providing access to historic information.

MemReread-4B improves on all eight tasks, including gains of 28.00\% on single-field retrieval and 23.50\% on numerical extrema.
Its improvements span field retrieval, retrieval across records, and aggregation.
ReMemR1-7B exhibits a different pattern: its largest gains occur in conditional counting and frequent-value identification, at 8.75\% and 3.96\%, while changes in tasks requiring multiple fields or records are smaller.
InfMem-4B improves on seven tasks but loses 1.75\% on frequent-value identification.
These results show that both the distribution and magnitude of task-level gains depend on the underlying memory agent.

\subsection{Component Ablation}
\label{subsec:component_ablation}

We assess component contributions by introducing them sequentially.
Starting from Flat, Tree adds structural sectioning, Tree+Path adds structural paths, and TreeAgent adds residual memory links with span retrieval.
Table~\ref{tab:main_results} reports all four variants using the same test examples and evaluation protocol.
Each comparison measures the incremental gain over the preceding variant.

\paragraph{Incremental contributions.}
Averaging equally across the four agents and four datasets, the mean answer score increases from 39.49\% for Flat to 41.35\% for Tree, followed by 41.70\% for Tree+Path and 43.10\% for TreeAgent.
Structural sectioning provides a substantial initial improvement, and adding structural paths further improves performance.
Once both components are in place, residual memory links with span retrieval yield additional gains in 15 of the 16 comparisons across agents and datasets.
These improvements show that access to historic memory provides value beyond organizing the current input by structure.

\paragraph{Variation in component gains.}
Sequential ablations on the JSON evaluation set demonstrate the benefits
of different TreeAgent components across agents.
Structural sectioning alone improves MemReread-4B by 13.04\%, raising
the mean answer score from 24.07\% to 37.11\%.
Structural paths and residual memory access provide further improvements.
For InfMem-4B, adding residual memory links with span retrieval to
Tree+Path raises the score from 58.15\% to 65.12\%, a further gain
of 6.97\%.
The results highlight the value of preserving local structure and
retrieving earlier evidence during memory updates.
TreeAgent integrates both mechanisms into existing agent workflows
without additional training.

\subsection{Budget Matching and Computational Overhead}
\label{subsec:budget-overhead}

The theoretical analysis of segment counts in Section~\ref{sec:structural-sectioning} informs the adjustment of the token budget to bring TreeAgent's segment count close to Flat's.
On the JSON evaluation set, the proportion of examples with identical segment counts under TreeAgent and Flat averages 74.8\% across the four agents.

We additionally measure model calls and cumulative input tokens over the complete inference process.
Across the four agents, TreeAgent increases the mean number of model calls by 3.7\%--7.6\% relative to Flat, while changes in cumulative input tokens range from $-5.7\%$ to $+4.0\%$.
On this evaluation set, TreeAgent improves performance with a modest
increase in model calls, while its cumulative input token usage remains
close to that of Flat.

\section{Related Work}

\paragraph{Memory agents.}
Recurrent Memory Transformer passes learned memory tokens across text segments to maintain information across processing steps \citep{bulatov2022recurrentmemorytransformer}.
At the agent level, MemAgent processes long inputs segment by segment and updates bounded memory before producing an answer \citep{yu2025memagentreshapinglongcontextllm}.
Subsequent work extends memory maintenance and access through gating for updates and stopping, retrieval, memory-guided rereading, and operations over memory banks \citep{sheng2026whentomemorizewhentostop,zhang2026learningtorememberendtoendtraining}.
InfMem incorporates retrieval into memory control \citep{wang2026infmemlearningsystem2memorycontrol}, MemReread uses existing memory to guide rereading \citep{ji2026memreread}, and ReMemR1 enables callbacks to historic memory states during memory updates \citep{shi2025lookbackreasonforward}.
TreeAgent complements these mechanisms by using the input hierarchy to organize the segments and historic evidence available at each update, while retaining the agent's memory update and answer procedures.

\paragraph{Structured document understanding.}
Document structure has been used to improve content representations and reading comprehension.
MarkupLM encodes XPath information in document representations \citep{li2022markuplm}, while DOM-LM and WebFormer also exploit webpage structure to model relationships among content elements \citep{deng2022domlmlearninggeneralizablerepresentations,wang2022webformer}.
WebSRC studies the roles of textual semantics and structural information through question-answering tasks that require understanding webpage structure \citep{chen2021websrc}.
TreeAgent examines how this structure can support the processing of long inputs segment by segment.
It preserves complete subtrees that fit the budget so that locally related fields are available within a single memory update, and attaches structural paths to identify each segment's hierarchical position when it is presented separately from the full input.

\paragraph{Hierarchical retrieval and reasoning.}
Hierarchical structures also organize retrieval and reasoning over long documents.
RAPTOR constructs retrieval trees through recursive clustering and summarization \citep{sarthi2024raptor}, while Chen et al. use discourse relations for hierarchical retrieval \citep{chen-etal-2026-beyond-chunking}.
ToM exploits document hierarchies to generate rationales at child nodes and aggregate them at parent nodes for bottom-up reasoning \citep{guo-etal-2025-tom}.
TreeAgent likewise uses hierarchical relations to organize information integration, operating on cumulative memory states formed during segment-wise processing.
Parent--child relations in the input tree link the historic state associated with a child to the final segment of its parent.
The structural relations determine which historic states are accessible and when, while the question guides the selection of relevant evidence within those states.
Information retained in historic states can therefore contribute again
to reasoning at higher structural levels without having to survive
every intervening memory update.

\section{Conclusion}
We presented TreeAgent for processing long hierarchical inputs using
recurrent memory agents.
TreeAgent uses the input hierarchy to organize sections, identify their
structural locations, and retrieve relevant evidence from historic memory
states, while retaining the underlying agent's memory update and answer
procedures.
Our theoretical analysis establishes upper bounds on segment counts and
guides the selection of a token budget to keep these counts close to those
of fixed-length segmentation.
Residual links allow relevant historic evidence to bypass intermediate
memory updates and inform reasoning at higher structural levels.
We evaluated TreeAgent on retrieval and aggregation tasks across four
formats and input lengths from 8K to 1M tokens.
Without additional training, TreeAgent improves the average score across
datasets for all four evaluated agents, with an overall gain of 3.61\%.
Sequential ablations show additional average gains from structural paths
and residual memory links with retrieval beyond those from sectioning alone.
Together, these findings support using input hierarchy to guide both the
organization of current observations and access to historic memory when
reasoning over long structured data. 

\bibliographystyle{ACM-Reference-Format}
\bibliography{main}

\appendix
\setcounter{equation}{0}
\numberwithin{equation}{section}
\renewcommand{\[}{\begin{equation}}
\renewcommand{\]}{\end{equation}}
\section{Additional Method Details}
\label{sec:supplementary-overview}
\label{sec:extended-method}

Flat and TreeAgent share the recurrent interface:
\[m_t=\mathcal{U}_{\mathcal{A},\theta}(q,m_{t-1},o_t).\]
The following details specify TreeAgent's observations and the controlled evaluation.

\subsection{Leaf Handling and Token Accounting}
\label{sec:leaf-handling}
For a divisible leaf exceeding $B$, \textsc{SplitText} produces consecutive
spans within $B$ at format-permitted boundaries. JSON strings retain escape
sequences and surrogate pairs; fragments restore quotation marks and enclosing
structure. Natural text boundaries are preferred when they sufficiently fill
the budget. A split leaf completes only after its final fragment.
YAML block scalars, undecomposed JavaScript expressions, interpolated templates,
and embedded HTML script or style content remain indivisible and may exceed
$B$. The budget counts native subtree or sibling-group serialization;
ancestor wrappers, paths, retrieved memory, and the agent's existing context
are accounted for separately.

\subsection{Budget Search}
\label{sec:budget-search}
Let $N=T_{\mathrm{flat}}(x,C)$ and $n(B)=|\mathcal{S}_B(x)|$.
The search evaluates integer budgets in $[C,2C-1]$ using the agent's tokenizer
and caches section counts. Candidate selection uses the lexicographic key
\[
\kappa(B)=\bigl(|n(B)-N|,\mathbb{I}[n(B)<N],B\bigr).
\]
Equal count differences favor a count not below $N$, then a smaller budget. Retain $C$ if $n(C)\leq N$.
Otherwise, evaluate $n(2C-1)$ and binary search for the transition to
$n(B)\leq\max\{N,n(2C-1)\}$.
If $n(2C-1)>N$, retain the final search budget; otherwise compare it with
its predecessor using $\kappa$.

\subsection{Paths, Residual Links, and Span Retrieval}

JSON paths start at \texttt{\$}; identifier keys use dot notation
and other keys use quoted brackets. Array indices use brackets.
For a section grouping consecutive children $u_i,\ldots,u_j$
of a parent $p$, the structural annotation is
$a_t=(\pi(p),[i,j])$, where both endpoints are included.
For example, the parent path \texttt{\$.records} together with
the range $[1,2]$ identifies the children at indices 1 and 2.
A text span split from a leaf retains that leaf's path.

The original tree and the section boundaries determine
$\ell(v)$, the index of the final section containing source
content from the subtree rooted at $v$. For a split leaf,
this is the section containing its final text span.
After every section $s_t$ is processed, the updated cumulative
memory $m_t$ is saved under the section index $t$.
Each node $v$ is associated with the saved state $m_{\ell(v)}$.
Nodes with the same final section therefore refer to the
same saved state.

Residual retrieval follows the parent--child links defined
in the main paper. A child $v$ of parent $p$ contributes the
source index $\ell(v)$ only at the target step $t=\ell(p)$,
and only if $\ell(v)<t$. The root is included as a possible
target parent. When multiple parents have the same final
section, their source indices are combined into
$\mathcal L_t$ and deduplicated before retrieval.
All source states are read from $H_{t-1}$; the current state
$m_t$ is saved only after the memory update.

For each saved state indexed by $\mathcal L_t$, divide its
text at line and sentence boundaries.
Lowercase terms, split camel case and underscores,
and remove a fixed set of function words. With term counts
$c_q,c_z$ and shared-term set $V(q,z)$, rank a span $z$ by
\[
\operatorname{score}(z,q)
=
\sum_{w\in V(q,z)}\min(c_q(w),c_z(w))
+0.25|V(q,z)|.
\]
Discard zero-overlap spans and resolve ties by source segment
index and span order within each saved state. Select at most four
spans across the combined candidate set, truncate each to 96 tokens,
and concatenate them to form $r_t$.
If $\mathcal L_t$ is empty or no candidate has positive
overlap, set $r_t=\emptyset$.

If each saved state contains at most $M$ tokens, storing the
states from $T$ sections requires $O(TM)$ tokens, in addition
to structural metadata. At step $t$, span extraction and
lexical scoring examine only the states indexed by
$\mathcal L_t$.

\subsection{Structural Sectioning Proof Details}
\label{sec:sectioning-proof-details}

We use the additive integer lengths and feasible partitions defined in
Section~\ref{sec:structural-sectioning}. For a node $v$, let $Q_B(v)$ be the minimum number of feasible
sections covering its subtree. A fitting subtree has $Q_B(v)=1$.
For an oversized node, let $\mathcal O_B(v)$ contain its children with
length greater than $B$, and let $\mathcal G_B(v)$ contain the maximal
nonempty consecutive sequences of its other children.
For $g\in\mathcal G_B(v)$, let $D_B(g)$ be the minimum number of sections
covering $g$ without splitting its child subtrees. Then
\begin{equation}
Q_B(v)=
\begin{cases}
1, & w(v)\le B,\\[2pt]
\displaystyle\sum_{u\in\mathcal O_B(v)}Q_B(u)
+\displaystyle\sum_{g\in\mathcal G_B(v)}D_B(g), & w(v)>B.
\end{cases}
\label{eq:tree-optimality-recurrence}
\end{equation}

To justify the recurrence, first consider a fitting child that a feasible
partition splits. Every section containing part of that child lies inside
it: a section extending outside would have to contain the whole child as
part of a sibling group. Replacing the internal sections by the whole child
therefore preserves feasibility and cannot increase the count. Repeating
this replacement yields an optimal partition that retains every fitting
child. No feasible section can contain an entire oversized child, or cross from
one side of that child to the other without containing it. Its internal
partition and the adjacent sibling sequences are consequently independent.
Their optima give a lower bound on any feasible partition, and concatenating
their optimal sections attains that bound.

Within each sequence $g$, deleting a prefix from a feasible partition leaves
a feasible partition of the suffix after empty sections are removed.
This proves the suffix monotonicity used in
Equation~\eqref{eq:sibling-optimality}. Greedy grouping attains $D_B(g)$,
and induction on tree height shows that Algorithm~\ref{alg:tree} attains
every term in Equation~\eqref{eq:tree-optimality-recurrence}.

For Corollary~\ref{cor:recursive-section-count}, collect the nonempty
sequences $\mathcal G_B(v)$ encountered during recursion in document order.
They partition the leaf units into $G_B$ groups and hence have $F_B=G_B-1$
boundaries between groups. If the root fits, use $G_B=1$ and $F_B=0$.
Writing $n_g$ for the number of output sections in group $g$ gives
\[
\sum_{g=1}^{G_B}(n_g-1)=N_B-G_B=N_B-1-F_B.
\]
These are the boundaries to which the packing inequality applies.
All section lengths are positive integers, so the two endpoint contributions
give the same subtraction of at least two as in
Equation~\eqref{eq:packing-boundary-sum}.

\section{Benchmark and Data Appendix}

The main paper defines the shared tasks, answer contract, and evaluation grid.
This appendix provides the additional information needed to inspect the
benchmark data: source provenance and licenses, parser units, construction and
validation procedures, and dataset inventory. Representative native-format excerpts are provided in our code repository.

\subsection{Benchmark Sources, Construction, and Validation}

We selected source collections whose native syntax exposes an explicit
hierarchy and whose visible records support deterministic lookup and
aggregation. The four benchmarks therefore test the same capabilities without
converting the original JSON, HTML, YAML, or JavaScript syntax into a common
representation.

\paragraph{JSON dataset.}

The JSON dataset contains API responses from seven source groups.
USGS provides earthquake events and their attributes
\citep{usgs_earthquake_catalog}; NVD provides CVE vulnerability records,
scores, and references \citep{nist_nvd}; and NIH provides research funding
projects through RePORTER and clinical trial records through ClinicalTrials.gov
\citep{nih_reporter_api,nih_clinicaltrials_api}.
GBIF provides species occurrence records and taxonomic information
\citep{gbif_api}, while Crossref provides metadata on scholarly works,
authors, and references \citep{crossref_rest_api}.
Contracts Finder provides UK procurement and contract records
\citep{contracts_finder}, and USAspending provides US federal contract
spending records \citep{usaspending_api}.

The parser unit is a JSON value inside an ordered response list.
Construction selects supporting records, generates a question in natural
language and a reference specification, and adds distractor
payloads. Validation executes the reference specification over the complete
visible context and checks answer uniqueness, with aggregation restricted
to the scope specified by the question. The 32K training and development
splits are used for model development; the test split, which contains all
eight length buckets, is held out. Multimodal clinical reasoning is outside this benchmark
\citep{hao2025surgical,HaoPen_Toward_MICCAI2026}.

\paragraph{HTML dataset.}

The HTML dataset uses pages from the WebSRC v1.0 test archive
\citep{chen2021websrc}, which declares CC BY 4.0. Its parser units are HTML
tables, lists, and DOM subtrees; text, element order, and native markup remain
visible to the model. Construction derives lookup and aggregation questions
from records with unique visible descriptions and fills the target length with
additional WebSRC pages.
Independent validation reparses every context, executes the stored reference
specification, checks source identifiers and answer uniqueness, and verifies
the length constraints under the tokenizers used during construction. The
dataset contains only test examples.

\paragraph{YAML dataset.}

The YAML dataset uses OpenAPI documents indexed by the APIs.guru directory
\citep{apisguru_openapi_directory}. For each document, we record its source
identifier and available license metadata. A parser unit is one YAML
OpenAPI document; paths, operations, parameters, and responses supply
records. Canonical documents are deduplicated before construction. Lookup examples
preserve their target fields and supporting documents across lengths, whereas
aggregation answers are recomputed after distractor documents are added. An
independent validator reparses the YAML, replays the document paths and
reference specifications, and checks source composition and context lengths.
The dataset contains only test examples.

\paragraph{JavaScript dataset.}

The JavaScript dataset evaluates static structured state written in JavaScript syntax; it
does not evaluate program execution or general program understanding. The
dataset is built from statically extractable object and array initializers in
six repositories distributed under the MIT License: ESLint, webpack, three.js,
Plotly.js, Babel, and Lodash
\citep{eslint_repository,webpack_repository,threejs_repository,plotlyjs_repository,babel_repository,lodash_repository}.
Calls, functions, spreads, identifier references, computed keys, parse errors,
sensitive strings, and ambiguous records are rejected. Every accepted
initializer is checked by a Tree-sitter parser and an independent JSON5
interpretation, which must produce the same static value without executing
source code. Each initializer retains its source repository, license, and
location. The dataset contains only test examples and measures static record
retrieval and aggregation.

\subsection{Dataset Inventory}

The four benchmarks contain 29,200 unique examples: 19,600 for the JSON dataset,
including its training and development splits, and 3,200 test examples for
each of the HTML dataset, the YAML dataset, and the JavaScript dataset. The evaluation in the main text uses
the 3,200 held-out test examples from each benchmark.

\section{Experimental Protocol}
\label{sec:experimental-protocol}

This section specifies the additional settings for the comparison without
additional training. It describes the shared controls, prompts and limits,
decoding procedure, and evaluation without changing the scope defined in the
main paper.

\subsection{Experimental Scope, Models, and Tokenizers}

The Cartesian product of four benchmarks, eight tasks, eight context lengths,
and 50 examples contains 12,800 distinct test examples. Evaluating every
example under Flat and TreeAgent with each of four memory agents produces
102,400 scored outputs. Each condition is evaluated once, and model parameters
remain fixed throughout this comparison.

We use the public checkpoints evaluated in the main paper, pairing each model
with its distributed tokenizer. Context length labels identify construction
buckets and do not imply identical token counts under the four tokenizers.
Section construction and token measurements use the tokenizer of the
evaluated model.

\subsection{Controlled Comparison}

Within each memory agent, Flat and TreeAgent use the same examples, checkpoint,
tokenizer, memory update procedure, available control actions, answer
interface, decoding parameters, and evaluator. Flat partitions the serialized
input into consecutive sections with a base budget of 5,000 tokens. TreeAgent
uses the same base budget and searches candidate integer budgets from 5,000
to 9,999 tokens as described in Appendix~\ref{sec:budget-search}.

During recurrent reading, Flat supplies the current text
section and the context constructed by the original memory
agent. TreeAgent supplies a structural section, its structural
path, and query-relevant spans retrieved from the saved states
indexed by $\mathcal L_t$, while retaining the original
agent's context and control procedure. The source indices
follow the residual links defined in the main paper:
a child state associated with $\ell(v)$ is eligible at
$t=\ell(p)$ only when $\ell(v)<t$.
At most four spans are supplied per update, each limited to
96 tokens. Retrieval precedes the memory update, and the
updated state is saved under the current section index.

\subsection{Prompts, Limits, Decoding, and Randomness}

Flat and TreeAgent receive the same underlying benchmark example and question.
The task instruction requests only the answer values and specifies whether
several values must follow the order stated in the question. It reveals no
schema, response index, structural path, supporting location, or reference
answer. Each memory agent retains the prompts and intermediate actions of its
released workflow. TreeAgent changes the reading observation described above
and adds no demonstrations specific to a task. Adaptive jailbreak robustness and output harmfulness remain untested
\citep{qi2026majic,qi2026darwin,zeng2026trace,xiu2026dynamicjailbreakingattack,huang2025dualbreachefficientdualjailbreakingtargetdriven,yang2026harmmetricevalbenchmarkingmetrics}.

Table~\ref{tab:generation-limits} reports the maximum number of generated
tokens for each agent's calls. The controller column covers explicit
planning, decomposition, retrieval, or callback actions when the workflow
uses them. MemAgent uses memory and final answer calls without a separate
controller. Request limits include both the prompt and the maximum generation.
They are 32,768 tokens for MemAgent and InfMem, and 40,960 tokens for
MemReread and ReMemR1.

\begin{table}[t]
\centering
\caption{Maximum generated tokens for agent calls. MemAgent does not use a
separate controller limit.}
\label{tab:generation-limits}
\begin{tabular}{lrrr}
\toprule
Memory agent & Memory & Answer & Controller \\
\midrule
MemAgent-7B  & 1,024 & 1,024 & n/a \\
InfMem-4B    & 1,024 & 1,024 & 1,024 \\
MemReread-4B & 1,024 & 1,024 & 8,192 \\
ReMemR1-7B   & 2,048 & 2,048 & 1,024 \\
\bottomrule
\end{tabular}

\end{table}

We use greedy decoding with temperature 0 and \texttt{top\_p}=1.0. The
inference seed is 42 when the serving interface accepts a seed. Dataset
sampling also uses seed 42, and deterministic preprocessing fixes the Python
hash seed to zero. 
The search bounds and selection rule were fixed for the complete test matrix
and were not tuned separately by benchmark, task, or context length. 
The selected budget was then computed deterministically for each example from its
input, tokenizer, and native format.

\subsection{Evaluation and Aggregation}

The primary metric is the mean answer score defined in the main paper. The
evaluator first extracts the final answer containing only the requested values
and then applies the rule for the corresponding task. Matching respects the
requested order for ordered retrieval tasks and ignores order for the
frequency ranking task. Text matching is insensitive to case and surrounding
formatting. Numerical matching allows a tolerance equal to the larger of 0.01
and 0.1\% of the reference magnitude.

An unavailable model response is represented by an empty extracted answer;
empty or malformed extracted answers receive zero.
Scores are first averaged over the 50 examples in each combination of
benchmark, task, context length, memory agent, and method. The score for each
benchmark averages the eight tasks and eight lengths equally. The reported
average over benchmarks gives equal weight to the JSON dataset, the HTML dataset, the YAML dataset,
and the JavaScript dataset. Thus, the aggregate is not weighted by context tokens or by the
number of model calls.

\label{sec:extended-results}
\subsection{Realized Computation and Paired Analysis}
\label{sec:realized-computation}
On the JSON dataset, section counts match for 74.8\% of examples. 
Table~\ref{tab:jsonruler_cost} reports realized usage. Paired 95\% confidence
intervals use the normal approximation over 3,200 matched examples per agent
(Table~\ref{tab:json-paired}); all exclude zero.

\begin{table}[t]
\caption{Realized computation on the JSON dataset: ratios of cumulative TreeAgent to Flat quantities. Input tokens include complete model inputs over all calls.}
\label{tab:jsonruler_cost}
\centering
\begin{tabular}{@{}lrr@{}}
\toprule
Memory agent & Model calls & Input tokens \\
\midrule
MemAgent-7B & 1.074 & 1.038 \\
InfMem-4B & 1.037 & 0.945 \\
MemReread-4B & 1.068 & 0.943 \\
ReMemR1-7B & 1.076 & 1.040 \\
\bottomrule
\end{tabular}
\end{table}

\begin{table}[t]
\caption{Paired score differences on the JSON dataset (\%) and 95\% confidence intervals. Absolute scores appear in Table~\ref{tab:main_results}.}
\label{tab:json-paired}
\centering
\begin{tabular}{@{}lrr@{}}
\toprule
Memory agent & Difference & 95\% CI \\
\midrule
MemAgent-7B & 1.80 & [0.62, 2.98] \\
InfMem-4B & 7.45 & [5.98, 8.91] \\
MemReread-4B & 15.50 & [13.77, 17.23] \\
ReMemR1-7B & 1.64 & [0.39, 2.89] \\
\bottomrule
\end{tabular}
\end{table}

\section{Exploratory Fine-Tuning}
\label{sec:treeagent-4b}

We conduct an exploratory experiment to assess whether additional training
improves InfMem-4B when used with TreeAgent. Starting from the released
checkpoint, we optimize the model with Group Relative Policy Optimization
(GRPO) \citep{shao2024deepseekmathpushinglimitsmathematical,hao2025surgery} on 16,000 JSON
training examples at 32K tokens. We select the checkpoint with the highest
mean answer score on a separate development set of 400 examples at 32K.
The selected checkpoint and the checkpoint before training are evaluated on
the same JSON test set, using the same TreeAgent observation process,
greedy decoding, and answer scoring.

Training runs for 30 steps with 64 prompts per batch and four rollouts per
prompt. We use a learning rate of $5\times10^{-6}$ and apply LoRA to all linear
modules with rank 32, scaling factor 32, and zero dropout. Training examples
are shuffled with seed 42, and checkpoints are saved every five steps.
The reward combines the answer score with two auxiliary terms of 0.05 each
for a valid interaction and completion of the memory update. Safety-aware fine-tuning remains future work
\citep{qi2026towards,wu2026datashield,qi2026ascentfirstorderoptimalfinetuning}.

Table~\ref{tab:treeagent-4b-length} reports the results at each input length.
The mean answer score increases from $65.117\%$ to $66.318\%$, an absolute
score difference of $1.201\%$. Scores improve at seven of the eight lengths,
with the largest difference of $4.458\%$ at 512K.

\begin{table}[!ht]
\centering
\caption{Mean answer scores (\%) on the JSON test set for InfMem-4B with
TreeAgent. Fixed and Trained denote the checkpoints before and after
additional training, respectively. Diff. denotes Trained minus Fixed.}
\label{tab:treeagent-4b-length}
\small
\setlength{\tabcolsep}{2.5pt}
\begin{tabular}{lrrr}
\toprule
Context & Fixed & Trained & Diff. \\
\midrule
8K   & 74.688 & 74.979 &  0.291 \\
16K  & 68.979 & 70.125 &  1.146 \\
32K  & 73.063 & 72.479 & -0.584 \\
64K  & 71.813 & 72.583 &  0.770 \\
128K & 65.625 & 66.583 &  0.958 \\
256K & 61.896 & 64.021 &  2.125 \\
512K & 51.188 & 55.646 &  4.458 \\
1M   & 53.688 & 54.125 &  0.437 \\
\midrule
Overall & 65.117 & 66.318 & 1.201 \\
\bottomrule
\end{tabular}

\end{table}

\FloatBarrier

\section{Evidence Scope and Statistical Interpretation}
\label{sec:limitations}
\vspace{0.5em}

The evaluation concerns recurrent memory agents on explicit hierarchies in JSON, HTML, YAML, and static JavaScript initializers, excluding visual authenticity
assessment \citep{li2026realnet,li2025toward,li2026synerdetect,zhang2025truemoe}. These four native formats provide complementary tests of record
binding, hierarchical lookup, and aggregation across sibling records. TreeAgent
uses the hierarchy exposed by the corresponding parser, so the comparison
focuses on how structured observations affect recurrent memory updates.

The comparison without additional training uses greedy decoding.
 The paired 
confidence intervals on the JSON dataset quantify variation
 in the score difference between TreeAgent and the Flat baseline 
over
 matched examples, complementing the aggregate results across agents,
formats, tasks, and context lengths.

\FloatBarrier

\end{document}